%% file: main.tex
\documentclass[11pt,a4paper]{article}
\usepackage[utf8]{inputenc}
\usepackage{graphicx}
\usepackage{xcolor}
\definecolor{esablue}{RGB}{0,57,158}
\definecolor{esalightblue}{RGB}{0,155,219}
\definecolor{esared}{RGB}{150,1,54}
\usepackage[hidelinks]{hyperref}
\hypersetup{
  colorlinks   = true, 
  urlcolor     = esalightblue, 
  linkcolor    = esablue, 
  citecolor   = esablue 
}
\usepackage{amsmath}
\usepackage{amssymb}
\newcommand{\gomez}{G\'{o}mez}

\usepackage[british]{babel}
\usepackage{csquotes}

\usepackage[textsize=tiny,
backgroundcolor=white,
textcolor=blue,
linecolor=blue,
bordercolor=blue]{todonotes}
\usepackage[capitalise]{cleveref}

\usepackage{caption}
\usepackage{subcaption}
\usepackage{tabularx}
\usepackage{multirow}
\usepackage{tablefootnote}
\usepackage[flushleft]{threeparttable}

\usepackage{booktabs}

\usepackage{amsfonts}

\usepackage[
backend=biber,
citestyle=numeric-comp,
bibstyle=numeric,
minbibnames=2,%
maxbibnames=3,%
maxcitenames=3,%
style=numeric,
autocite=plain,
sortcites=true,
uniquename=init,
giveninits=true,
isbn=true,
doi=true,
url=true,%
natbib=true,%
urldate=short,
sorting=none,
arxiv=abs,
dateabbrev=false
]{biblatex}
\renewbibmacro{in:}{}

\AtEveryBibitem{%
	\ifentrytype{inproceedings}{
		\clearlist{publisher}
		\clearfield{volume}
		\clearfield{pages}
	}{}
	\clearlist{address}
	\clearlist{location}
    \ifentrytype{misc}{}{
        \clearfield{month}
    	\clearfield{day}
        \clearfield{url}
        \clearfield{urldate}
    }
	\clearfield{series}
	\clearfield{primaryclass}
	\clearfield{eprintclass}
}

\title{Selected  Trends in Artificial Intelligence for Space Applications}
\author{Dario Izzo\thanks{Advanced Concepts Team (ACT), European Space Research \& Technology Centre (ESTEC), Keplerlaan 1, 51014 AG Noordwijk (Netherland)} , Gabriele Meoni$^*$\thanks{$\Phi$-lab,  European Space Research Institute (ESRIN), Via Galileo Galilei, 1, 00044 Frascati RM (Italy)}, Pablo \gomez$^*$\\ Dominik Dold$^*$, Alexander Zoechbauer$^*$}
\date{}

\begin{document}

\maketitle
\begingroup
\hypersetup{linkcolor=black}
\endgroup
\input{sections/intro}

\input{sections/differentiable}
\input{sections/onboardAI}

\printbibliography


\end{document}

%% file: sections/intro.tex
\section{Introduction}
The development and adoption of artificial intelligence (AI) technologies in space applications is growing quickly as the consensus increases on the potential benefits introduced. 
As more and more aerospace engineers are becoming aware of new trends in AI, traditional approaches are revisited to consider the applications of emerging AI technologies. 
Already at the time of writing, the scope of AI-related activities across academia, the aerospace industry and space agencies is so wide that an in-depth review would not fit in these pages. 
In this chapter we focus instead on two main emerging trends we believe capture the most relevant and exciting activities in the field: differentiable intelligence and on-board machine learning. 
Differentiable intelligence, in a nutshell, refers to works making extensive use of automatic differentiation frameworks to learn the parameters of machine learning or related models. Onboard machine learning considers the problem of moving inference as well as learning of machine learning models onboard. 

Within these fields, we discuss a few selected projects originating from the European Space Agency's (ESA) Advanced Concepts Team (ACT), giving priority to advanced topics going beyond the transposition of established AI techniques and practices to the space domain, thus necessarily leaving out interesting activities with a possibly higher technology readiness level.
We start with the topic of differentiable intelligence by introducing Guidance and Control Networks (G\&CNets), Eclipse Networks (EclipseNETs), Neural Density Fields (geodesyNets) as well as the use of implicit representations to learn differentiable models for the shapes of asteroids and comets from LiDAR data. We then consider the differentiable intelligence approach in the context of inverse problems and showcase its potential in material science research. 
In the next section we investigate the issues that are introduced when porting generic machine learning algorithms to function onboard spacecraft and discuss current hardware trends and the consequences of their memory and power requirements. 
In this context we present the cases of the European Space Agency satellites $\Phi$-sat and OPS-SAT, introducing preliminary results obtained during the data-driven competition \lq\lq the OPS-SAT case\rq\rq\ on the real-time onboard classification of land use from optical satellite imagery.
All in all, we wish this chapter to contain a list of new results and ideas able to stimulate, in the coming years, the interest of practitioners and scientists and to inspire research directions further into the future.

%% file: sections/differentiable.tex
\section{Differentiable Intelligence}
The term \lq\lq differentiable intelligence\rq\rq\ or \lq\lq differential intelligence\rq\rq\ 
has been recently used to indicate, in general, learning algorithms that base their internal functioning on the use of differential information. 
Many of the most celebrated techniques in AI would not be as successful if they were not cleverly exploiting differentials. 
The backpropagation algorithm, at the center of learning in artificial neural networks (ANNs), leverages the first and (sometimes) second order derivatives of the loss in order to update the network parameters \cite{rumelhart1995backpropagation}. 
Gradient boosting techniques \cite{natekin2013gradient} make use of the negative gradients of a loss function to iteratively improve over some initial model. 
More recently, differentiable memory access operations \cite{graves2016hybrid, graves2014neural} were successfully implemented and shown to give rise to new and exciting neural architectures. 
Even in the area of evolutionary computations, mostly concerned with derivative-free methods, having the derivatives of the fitness function is immensely useful, to the extent that many derivative-free algorithms, in one way or another, seek to approximate such information (e.g. the covariance matrix in CMA-ES \cite{hansen2016cma} is an approximation of the inverse Hessian). 
At the very core of any differentiable intelligence algorithm lie \lq\lq automated differentiation\rq\rq\ techniques, whose efficient implementation determine, ultimately, the success or failure of a given approach. 
For this reason, in recent years, software frameworks for automated differentiation have been growing in popularity to the point that software frameworks such as JAX, PyTorch, TensorFlow and others \cite{van2018automatic} are now fundamental tools in AI research where lower order differentials are needed and an efficient implementation of the backward mode automated differentiation is thus of paramount importance. Higher order differential information, less popular at the time of this writing, is also accessible via the use of dedicated frameworks such as pyaudi \cite{pyaudi} or COSY infinity \cite{makino2006cosy}. 
In the aerospace engineering field, many of the algorithms used to design and operate complex spacecraft can now leverage these capabilities, as well as make use of the many innovations continuously introduced to learning pipelines. 
A number of applications deriving from these advances have been recently proposed and are being actively pursued by ESA's ACT: a selection is briefly introduced in the following sections.

\subsection{G\&CNets}
\begin{figure}[tb]
\centering
\includegraphics[width=\linewidth]{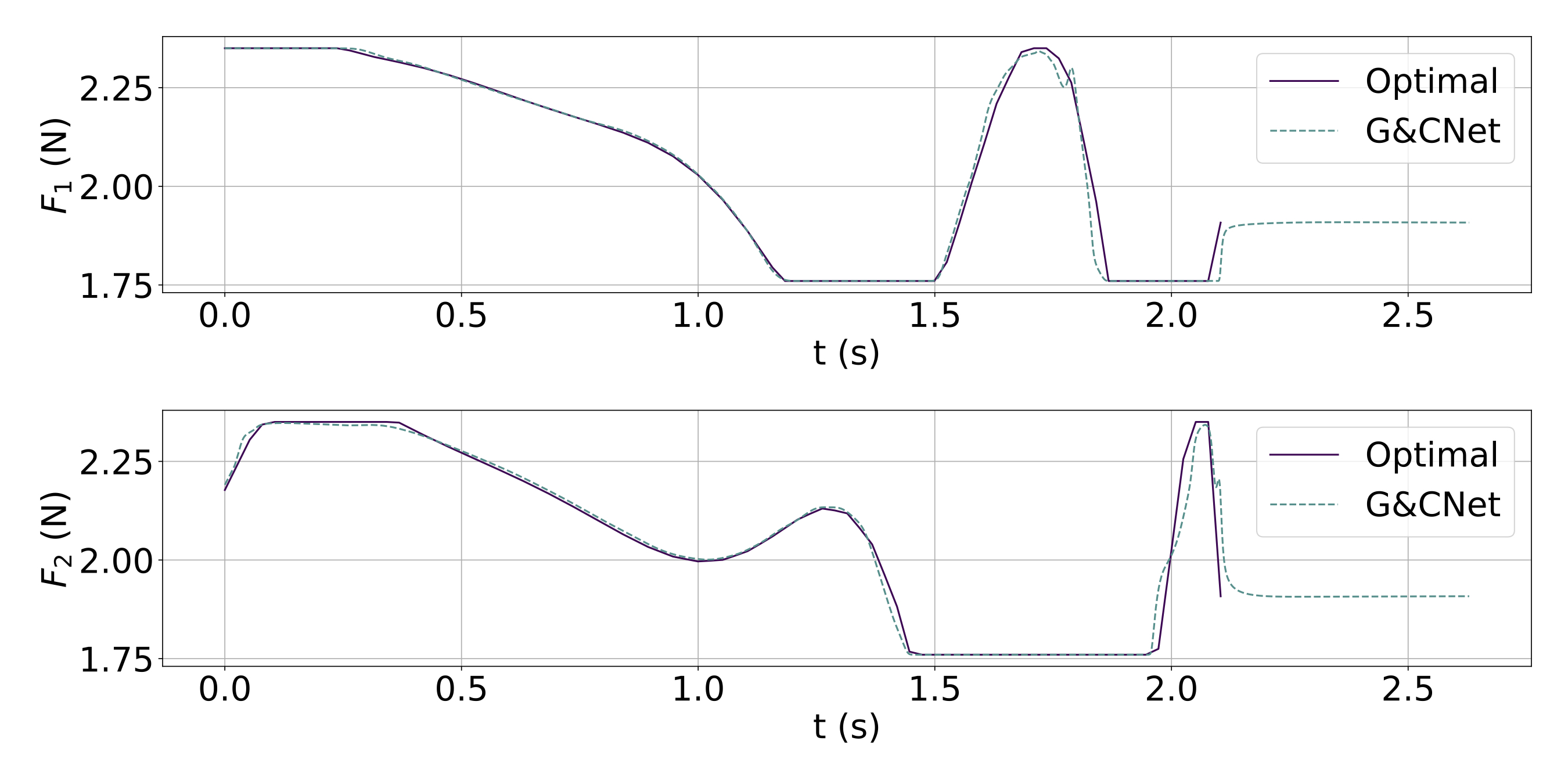}
\caption{History of the forces (called $F_1, F_2$) controlling a quadcopter during a simulated aggressive flight (i.e. with a strong emphasis on time optimality). The ground truth optimal solution is shown as well as the one produced using a G\&CNet. Reproduced from \cite{li2020aggressive}.\label{fig:gecnet}}
\end{figure}
The acronym G\&CNet \cite{sanchez2018real} stands for Guidance and Control Networks and was introduced to refer to a learning architecture recently proposed in the context of real-time optimal spacecraft landing. 
The basic idea is to go beyond a classical G\&C scheme, which separates the spacecraft guidance from its control, and have instead an end-to-end ANN able to send optimal commands directly to the actuators using only the state information computed from the navigation system as input. 
Indicating the spacecraft state with $\mathbf x$ and its control with $\mathbf u$, it is known how, under mild hypothesis, such a mapping exists (i.e. the optimal feedback $\mathbf u^*(\mathbf x)$) and is unique \cite{bardi1997optimal}. 
The structure of the optimal feedback $\mathbf u^*$ is, in general, that of a piece-wise continuous function, which suggests the possibility to use a neural network as its approximator. 
We are thus allowed to write $\mathbf u^* = \mathcal N_{\theta}(\mathbf x) + \epsilon$, where $\mathcal N$ is any neural model with parameters $\theta$ and $\epsilon$ an error that can, in principle, be sent to zero as per the Universal Approximation Theorem \cite{guhring2020expressivity}. 
We are now left with the task to find (learn) the values of $\theta$ as to make $\epsilon$ sufficiently small. 
The use of imitation learning has been shown, in this context, to be able to deliver satisfactory results in a variety of scenarios including spacecraft landing \cite{sanchez2018real}, interplanetary spacecraft trajectory design \cite{izzo2021real, izzo2022seb} as well as drone racing \cite{li2020aggressive}. 
In all cases, we first generate a large number of optimal flight profiles and we record them into a training dataset containing optimal state-action pairs $(\mathbf x_i, \mathbf u^*_i)$. We then use the dataset to learn the network parameters $\theta$ solving the resulting nonlinear regression problem (i.e. in a supervised learning approach). 
The problem of efficiently assembling the dataset to learn from is solved introducing an ad hoc technique called backward propagation of optimal samples \cite{izzo2021real, izzo2022seb}. At the end, the G\&CNet $\mathcal N_{\theta}(\mathbf x)$ is used in simulation, and later on real hardware, to produce optimal flights. 
In Fig.~\ref{fig:gecnet}, we reproduce a plot from \cite{li2020aggressive} where a G\&CNet is used to optimally control the two-dimensional dynamics $\mathbf f(\mathbf x, \mathbf u)$ of a quadcopter. Starting from some initial condition, the results from the following simulations are compared: $\dot {\mathbf x} = \mathbf f(\mathbf x, \mathbf u^*)$ (optimal) and $\dot {\mathbf x} = \mathbf f\big(\mathbf x, \mathcal N_\theta(\mathbf x)\big)$ (G\&CNet).
The figure compares the control history during both simulations and well illustrates, in this specific case, several generic aspects of the method: the satisfactory approximation accuracy, the stability of the resulting flight as well as the sensible behaviour once the optimal flight is over (in this case after $t^*\approx 2.1$ seconds). The stability of the system dynamics under the influence of a G\&CNet is studied in details in \cite{izzo2020stability}, while the learning pipeline optimization and tuning is discussed in general in \cite{tailor2019learning}. 

\subsection{Implicit representations of irregular bodies}
\begin{figure}[tb]
\centering
\includegraphics[width=\linewidth]{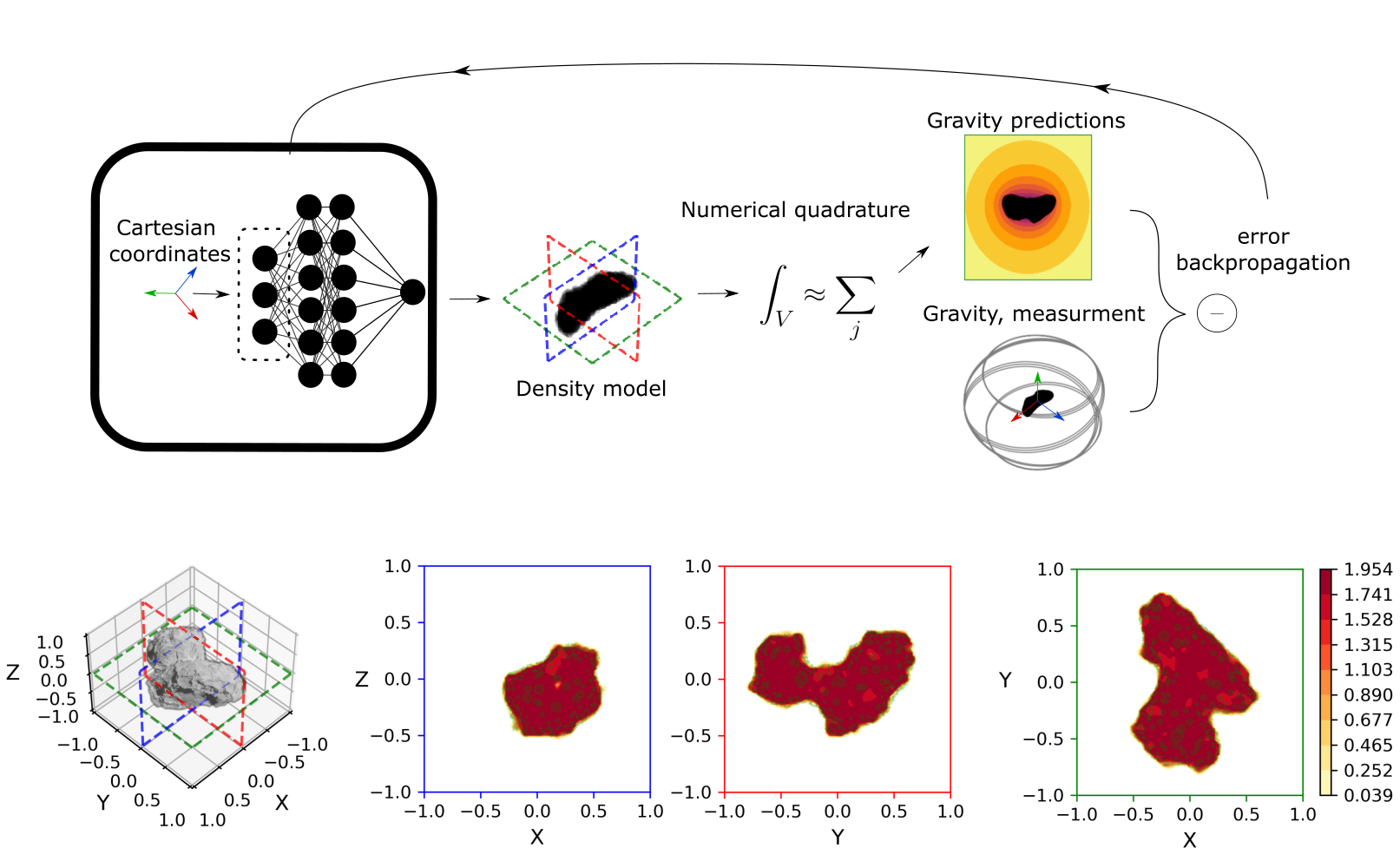}
\caption{Top: generic training setup for geodesyNets. Bottom: visualization of the learned neural density field in the specific case of a homogeneous model of the comet 67p/Churyumov-Gerasimenko. Reproduced from \cite{izzo2021geodesy}.\label{fig:67p}}
\end{figure}
The expressivity of ANNs, mathematically discussed in \cite{guhring2020expressivity}, found recently a remarkable showcase in \cite{mildenhall2021nerf} where an original machine learning pipeline was used to learn a so-called NEural Radiance Field (NERF): a relatively small, albeit deep, feedforward neural network mapping Cartesian coordinates ($x,y,z$) and ray direction ($\theta, \varphi$) to the local volume density and directional emitted color of a 3D scene (e.g. an object like a digger). 
After training, the network was able to render two-dimensional views of the target scene with unprecedented levels of details. We argue that an important take-away message from this work -- and the many that followed and introduced the term \emph{implicit representations} to refer to this type of networks -- is the ability of a relatively simple network architecture to encode highly complex objects with great precision in its parameters. 
The availability of a shape model of an irregular astronomical body is important when planning orbital maneuvers and close-proximity operations, but is also of great interest to scientists trying to reconstruct the complex history of the Solar System dynamics and understand its current state.
Following this remark, it is natural to ask whether the shapes of irregular bodies in the solar system, such as comets, asteroids and even spacecraft, could be a) represented with sufficient accuracy by an ANN and b) learned from measurements typically available in some space mission context. 
These scientific questions are at the core of several innovative projects the ACT is currently developing towards higher technology readiness levels.

\subsubsection{GeodesyNets} \label{GeodesyNets}
Any body orbiting in the Solar System can be described, for the purpose of modelling the resulting gravitational field, by its density $\rho(x,y,z)$. Such a function will be zero outside the body and discontinuous in its interior as to model possible fractures and material heterogeneity, and thus a great candidate for being represented by an ANN: a geodesyNet \cite{izzo2021geodesy}.
We thus set $\rho(x,y,z) = \mathcal N_\theta(x,y,z) + \epsilon$ and, again, learn the parameters $\theta$ as to make $\epsilon$ vanish. Since we do not have access to the actual values of the body's density we cannot set up a standard supervised learning pipeline.
Instead, we assume to be able to measure the gravitational acceleration $\mathbf a_i$ at a number of points $\mathbf X_i = [X_i,Y_i,Z_i]$, for example along a putative spacecraft orbit. At any of those points, the network will predict $\mathbf {\hat a}_i(\mathbf x_i) = - \iiint_V \frac{\mathcal N_\theta(x,y,z)}{r_i^3}\mathbf r_i dV$, where $\mathbf r_i = [X_i-x, Y_i-y,Z_i-z]$. The difference between the predicted and the measured acceleration can then be used to learn the network parameters $\theta$. As detailed in the work introducing geodesyNets, the final representation of the body density (i.e. the so-called neural density field) has several advantages over more classical approaches such as spherical harmonics, mascon models or polyhedral gravity models \cite{izzo2021geodesy}. Most notably it has no convergence issues, it maintains great accuracy next to the asteroid surface, it can be used to complement a shape model and propose plausible internal structures and it is differentiable. In Fig. \ref{fig:67p} the overall scheme and an example of the results obtained after training a GeodesyNet is shown. A model of the comet 67p/Churyumov-Gerasimenko is used to produce a synthetic scenario where to test the training. The comet is assumed to be perfectly homogeneous and all units are non-dimensional (see \cite{izzo2021geodesy} for details). The results reported here are qualitatively similar for other irregular bodies and the main take away is that geodesyNets are able to represent the body shape, its density and the resulting gravity field with great accuracy. Furthermore, the model parameters can be learned from the body's gravitational signature. 
An interesting question, addressed in recent works \cite{izzo2021geodesy, moritz}, is the behaviour of geodesyNets when the interior of the target irregular body is not uniform and the gravitational signature is noisy (for example being contaminated by non-gravitational effects). 
Under the additional assumption that a shape model has already been obtained for the body, it is possible to modify the geodesyNet pipeline as to have the network directly predict the deviation from a homogeneous density distribution, thus targeting a differential density $\partial\rho(x,y,z) = \rho - \rho_U = \mathcal N_\theta(x,y,z)$. 
In this case, internal details of the body's heterogeneous density become clear and sharper \cite{izzo2021geodesy}. 
The approach, in this case utilized to see inside the body, cannot, however, evade the constraints imposed by Newton's shell theorem (i.e. the impossibility to invert gravity fields uniquely) and the effects of measurement sensitivity and noise (i.e. the possibility to measure the gravitational signature of heterogeneity).

\subsubsection{EclipseNets}
During an eclipse event (e.g. when the Sun light gets obscured as an orbiting object enters the shadow cone of some other body) the orbital dynamics changes significantly, in particular for small area-to-mass ratio objects such as orbiting dust, i.e., debris pieces or pebbles detached from rotating asteroids. Due to the possible irregular shape of the eclipsing body as well as diffraction and penumbra effects, modelling these events with precision is in most cases computationally expensive.

\begin{figure}[tb]
\centering
\includegraphics[width=\linewidth]{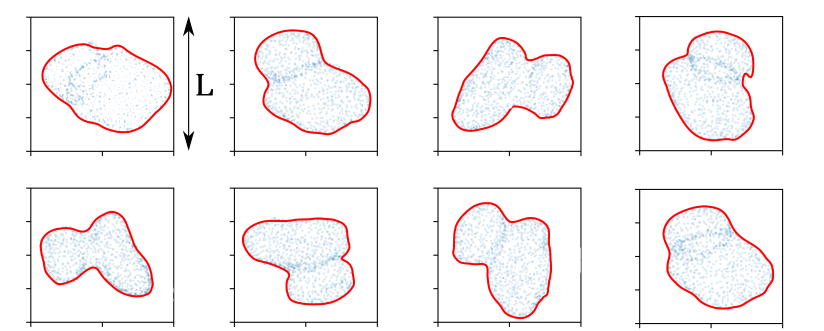}
\caption{Shadow cone predicted by an EclipseNet (red) for the comet 67p/Churyumov-Gerasimenko along eight different random directions. The vertices of the polyhedral model of the comet used to compute the ground truths are shown in blue. During training, non-dimensional units (i.e. $L$) are used. \label{fig:eclipsenet}}
\end{figure}

Ignoring diffraction and penumbra effects, the neural architecture EclipseNet introduced in \cite{biscani2022reliable} is a differentiable model capable of reproducing the complex shape of shadow cones.
An EclipseNet is a relatively simple feedforward neural architecture trained to predict the value of the complex and discontinuous eclipse function $\mathbf F(\mathbf r, \mathbf{\hat i}_S)$. 
The eclipse function maps the Cartesian position vector $\mathbf r$ and the light direction $\mathbf{\hat i}_S$ into a positive number if $\mathbf r$ is outside the shadow cone (cast along the direction $\mathbf{\hat i}_S$) and negative otherwise. 
The idea here is that the zero-level curves of such a function, once $\mathbf{\hat i}_S$ is fixed, determine the shape of the body's shadow cone. The concept is similar to that of the signed distance function (SDF) \cite{osher2003signed} used widely in image reconstruction, but is defined on a 2D projection and parameterized  by the direction of the light source.
One possible way to have such a function is to define it as the distance of $\mathbf r$ from the shadow cone (degenerated into a cylinder in this case) whenever $\mathbf r$ is outside the cone. 
The eclipse function can otherwise (when inside the shadow) be defined as the length of the ray portions inside the body. Both quantities are easily computed  using the Möller-Trumbore algorithm \cite{moller1997fast} starting from a polyhedral shape model of the body. Hence, a standard regression pipeline can be applied to learn the network parameters $\theta$ so that we can write $\mathbf F(\mathbf r, \mathbf{\hat i}_S) = \mathcal N_\theta(\mathbf r, \mathbf{\hat i}_S) + \epsilon$. In Fig.\ref{fig:eclipsenet}, we report the results obtained from training an EclipseNet in the case of the irregular shape of the comet 67p/Churyumov-Gerasimenko being the source of the eclipse. A feedforward neural network with 6 layers of 50 $\tanh$ units is used in this case for a dataset containing 10000 values of an eclipse function computed for 300 distinct views forming a Fibonacci spiral. 
The remarkable expressivity of such a relatively simple network can be clearly seen in the figure where the zero level curve of the network outputs is plotted for a fixed direction. The network thus appears to be able to represent well the shadow cone. 

To model the influence of eclipses over the orbiting dynamics, one can use a simple model. Assume a reference frame attached to the main body, which is uniformly rotating with an angular velocity $\boldsymbol \omega$. The solar radiation pressure is introduced via an acceleration $\eta$ acting against the Sun direction $\hat{\mathbf i}_S$ and regulated by an eclipse factor $\nu(\mathbf r) \in [0,1]$ determining the eclipse-light-penumbra regime.
In particular, $\nu=1$ when the body is fully illuminated and $\nu=0$ when fully eclipsed (umbra).  
Formally, the following set of differential equations are considered:
\begin{equation}
    \label{eq:eom_mascon}
\ddot {\mathbf r} = \mathbf a_g - 2 \boldsymbol\omega \times \mathbf v - \boldsymbol \omega \times\boldsymbol\omega \times \mathbf r - \eta \nu(\mathbf r) \hat{\mathbf i}_S(t)\,.
\end{equation}
To numerically solve these equations, one needs to know the value of $\nu(\mathbf r)$. To do so, one can detect the eclipse function's sign change as an event using the EclipseNet. This avoids more complex raytracing computations (e.g. running the Möller-Trumbore algorithm at each instance) as to determine whether the solar radiation pressure is active. This also avoids the need to store and access the body polyhedral model compressing that information into the network parameters. Furthermore, the differentiability of the EclipseNet model allows, in this use case, the usage of a guaranteed event detection scheme based on Taylor integration \ref{fig:eclipsenet}.

\subsubsection{Differentiable 3D shapes from LiDAR measurements}
Terrestrial-based observations are often used to determine the preliminary shape model of irregular bodies and use it during the design of a space mission targeting operations in its proximity. 
Such a model, used mainly for navigation purposes and sufficient for that purpose, is obviously inaccurate and gets refined at a later stage for scientific purposes -- thanks to the data produced by the various scientific payloads typically on board. 
One is often a LiDAR (LIght Detection And Ranging), a technique that can be used to measure surface features as well as to support scientific mapping and ultimately close operations. 
Such is the case, e.g., for the upcoming HERA mission \cite{michel2018hera} (ESA) as well as the past Hayabusa (JAXA) and Osirix-Rex (NASA) missions. 
In this context it is of interest to study whether AI-based techniques can be used to introduce new algorithms able to build a full 3D shape model directly from LiDAR data. Such algorithms, accelerated by edge computing solutions, can be considered to be run onboard and thus be able to leverage incoming LiDAR data to continuously improve the shape model available to the spacecraft. 
\noindent
With this vision in mind, and inspired by previous work in the field of computer vision, in particular \cite{osher2003signed, Peng2021SAP}, we developed and preliminary tested novel pipelines able to reconstruct a triangle mesh, representing the surface of the irregular astronomical body from a point cloud consisting of LiDAR points $\mathbf{x}_i$.
A first idea is to use neural implicit representations to learn the surface information encoded in the Signed Distance Function \cite{osher2003signed} computed from LiDAR data as: 
\begin{equation}
\Phi(\mathbf{x},\Omega) =
\begin{cases}
  \inf_{\mathbf{y} \in \partial\Omega} ||\mathbf{x}-\mathbf{y} ||^2 &\text{if } \mathbf{x}\in \Omega \,,\\
  -\inf_{\mathbf{y}  \in \partial\Omega} ||\mathbf{x}-\mathbf{y} ||^2 &\text{if } \mathbf{x}\in \Omega_c \,.
\end{cases}
\end{equation}

The SDF describes the minimum distance between any point $\mathbf{x}$ to the surface $\partial\Omega$ of a three-dimensional geometry $\Omega$. 
Notably the points where $\Phi(\mathbf{x})=0$ is the surface. It is then possible to train a feedforward neural network to predict $\Phi(\mathbf{x})$ knowing that $\Phi(\mathbf{x}_i)=0$ at the LiDAR points $\mathbf{x}_i$ and adding the condition $|\nabla \Phi(\mathbf{x})| = 1$ everywhere. A shape model can subsequently be obtained using a marching cubes algorithm, which constructs a mesh from the zero level set of $\Phi(\mathbf{x})$.
A second method studied emerged through recent advances in 3D scene reconstruction made by \cite{Peng2021SAP} where a Differential Poisson Solver (DPS) for surface reconstruction is introduced. The solver is used to compute the Indicator Function defined as:
\begin{equation}
\chi(\mathbf{x},\Omega) =
\begin{cases}
  0.5 &\text{if } \mathbf{x}\in \Omega / \partial\Omega \,,\\
  0 &\text{if } \mathbf{x}\in \partial\Omega \,,\\
  -0.5 &\text{if } \mathbf{x}\in \Omega_c \,.
\end{cases}
\label{ind:definition}
\end{equation}
The DPS solves the Poisson equation defined as $\nabla^2 \chi(\mathbf{x}) =  \nabla \cdot v(\mathbf{x})$ for $\chi(\mathbf{x})$, where $v(\mathbf{x})$ is a vector field of points and normal vectors of a predefined point cloud, by using spectral methods (Fast Fourier Transforms).
From the zero level set of $\chi(\mathbf{x})$, a mesh can be obtained using the marching cubes algorithm and subsequently comparing the Chamfer distance to the LiDAR data to inform a loss and thus update -- thanks to the differentiability of such a pipeline -- the shape model iteratively. 
In this second case, no intermediary ANN is used as the position of the vertices in the shape model are directly updated.
\begin{figure}
     \centering
     \begin{subfigure}[b]{0.32\textwidth}
         \centering
         \includegraphics[width=\textwidth]{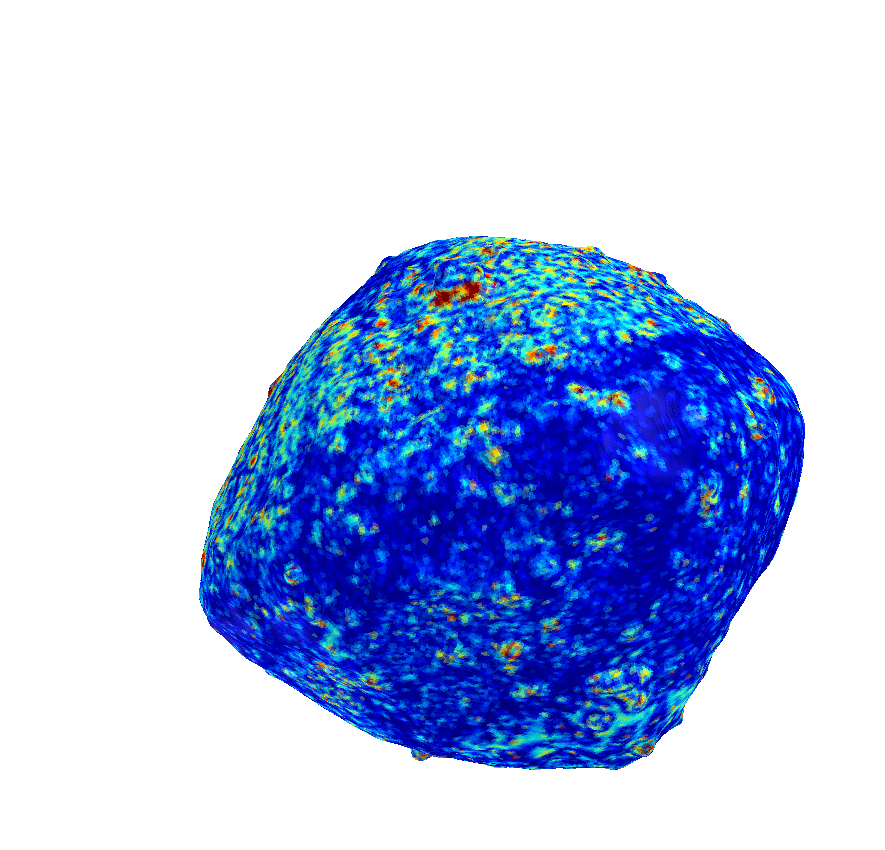}
         \caption{DPS}
     \end{subfigure}
     \hfill
     \begin{subfigure}[b]{0.32\textwidth}
         \centering
         \includegraphics[width=\textwidth]{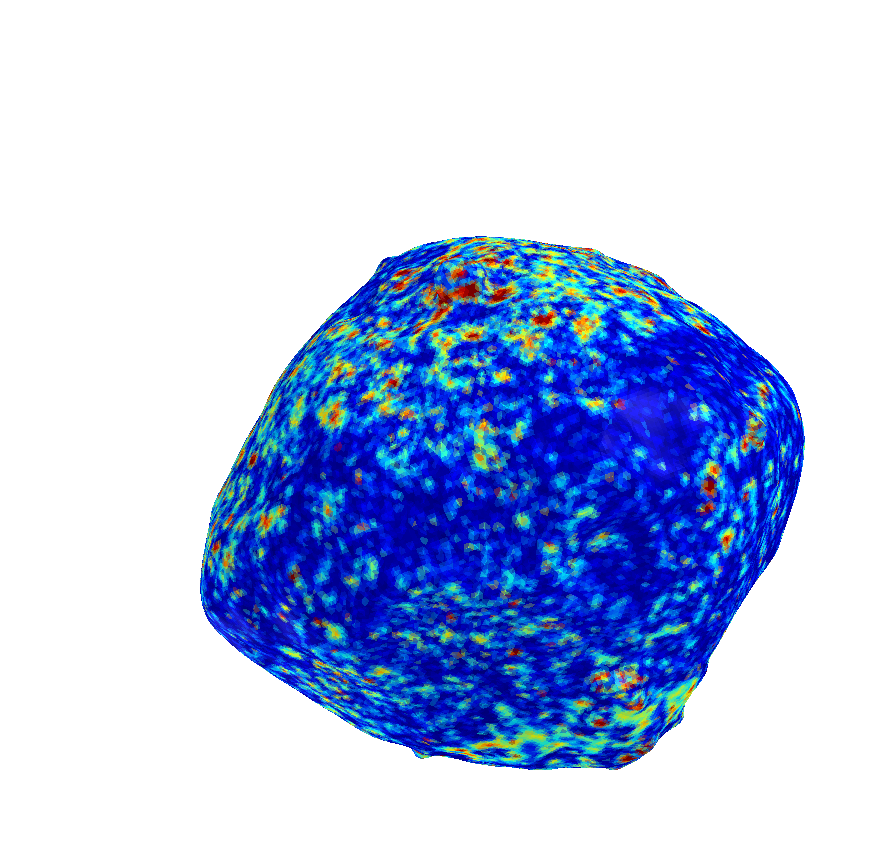}
         \caption{Implicit Representation}
     \end{subfigure}
     \hfill
     \begin{subfigure}[b]{0.32\textwidth}
         \centering
         \includegraphics[width=\textwidth]{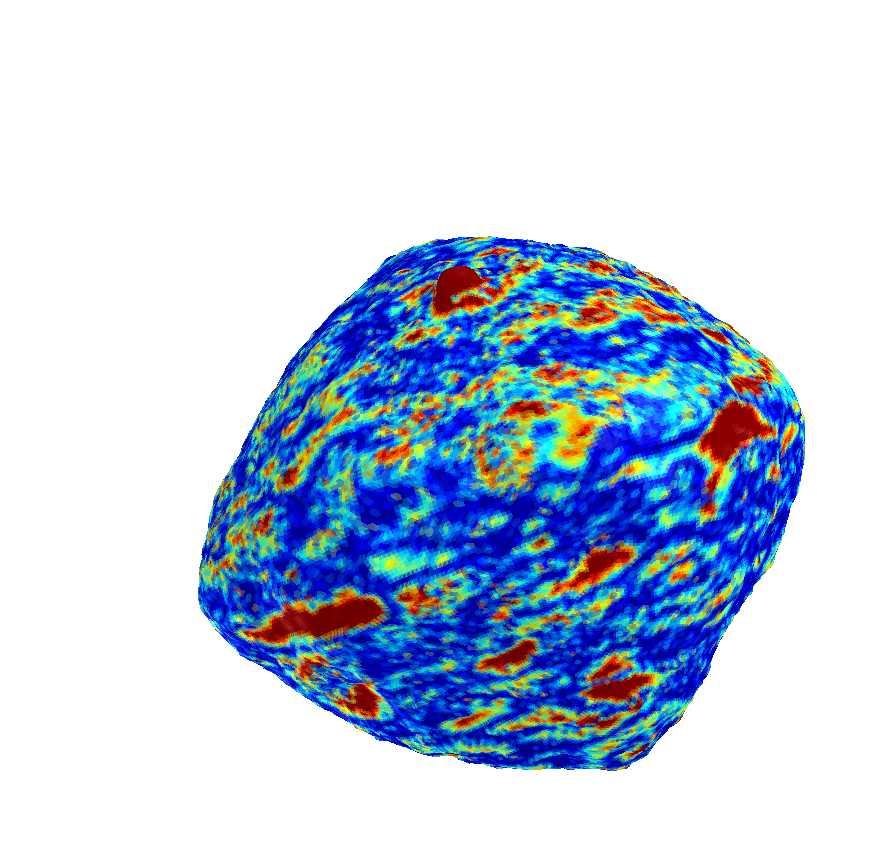}
         \caption{Shape model from images}
     \end{subfigure}
     \caption{Difference between the reconstructed shape models of the asteroid Bennu and the LiDAR ground truth. a), b) Shape model derived from the proposed differentiable methods. c) Shape model reconstructed from camera images (publicly available online from the Osirix-Rex mission). Regions with high/low error are encoded in red/blue.}
     \label{fig:mesh_comparison}
\end{figure}
We report in Fig.~\ref{fig:mesh_comparison} a visualization of the error one can obtain running these algorithm in the case of the LiDAR reading available for the asteroid Bennu from the Osirix-Rex mission. In the figure, the ability of the proposed method to construct an accurate body shape compatible with the LiDAR measurements appears evident in comparison to, for example, the shape model constructed using camera images.

\subsection{Inverse Material Design}

Modeling physical processes via differentiable code, involving also highly parameterized neural models, allows for the development of new powerful inversion algorithms as well as solvers. The approach has been recently applied to complex phenomena such as fluid mechanics \cite{cai2022physics,thuerey2021physics,gomez2018laryngeal} and more generally those defined by complex PDEs \cite{yang2021b}. In the ACT this is being explored also in the context of material design, e.g., \cite{gomez2022nidn} and \cite{dold2023lattice}, which will be discussed in the following sections.

\subsubsection{Neural Inverse Design of Nanostructures}

\begin{figure}[tb]
\centering
\includegraphics[width=\linewidth]{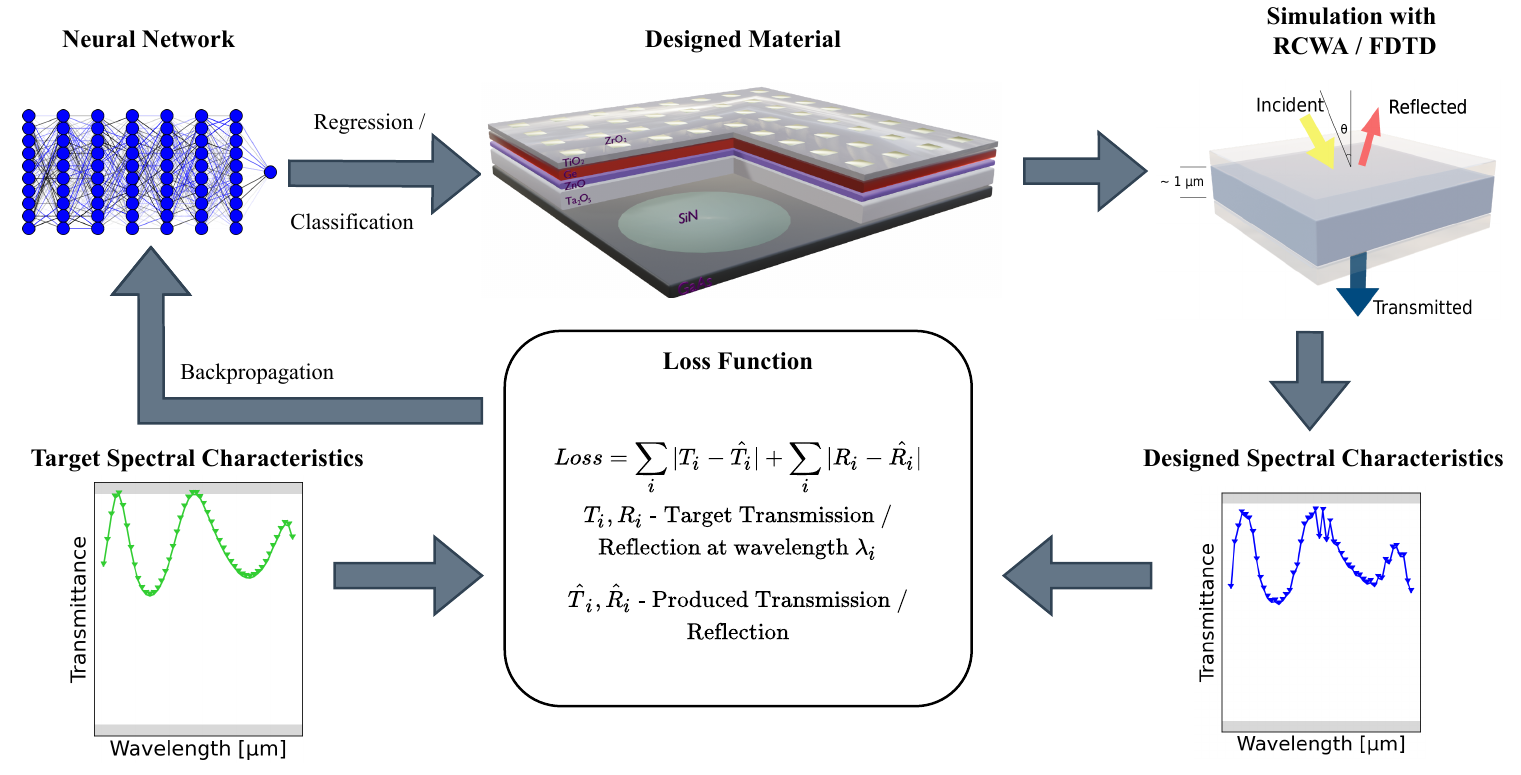}
\caption{Training setup for NIDN. \label{fig:nidn_1}}
\end{figure}

The open-source Python module called \textit{Neural Inverse Design of Nanostructures} (NIDN) has been released and allows for the  design of nanostructures targeting specific spectral characteristics (Absorptance, Reflectance, Transmittance). NIDN has applications, e.g., in designing  solar light reflectors, radiators to emit heat, or solar sails. Naturally, the applications expand beyond the space domain. In contrast to previous works \cite{Ma_Meta_Refl__VAE_2020,Nadell_Meta_Trans_CST_FFDS_2019} NIDN utilizes internal, differentiable Maxwell solvers to build a completely differentiable pipeline. Among other advantages, this eliminates the need for any form of training dataset and simplifies the inversion as the training process in NIDN aims to find the best solution for a specific set of spectral characteristics, and not to find a general inverse solution of the Maxwell equations. \\
On the inside, NIDN relies on two different solvers for the Maxwell equations depending on desired complexity and wavelength range. The first, TRCWA, is based on rigorous coupled-wave analysis (RCWA) \cite{moharam1981rigorous} and a highly efficient solver for stacked materials. The second one is a finite-difference time-domain (FDTD) \cite{sullivan2013electromagnetic} solver, which is computationally more expensive but a highly accurate method. A detailed description of the pipeline for designing a material with NIDN is given in Fig.~\ref{fig:nidn_1}. Fundamentally, the idea is to iteratively improve the material's permittivity based on the gradient signal backpropagated through the Maxwell equations solver.

\begin{figure}[tb]
\centering
\includegraphics[width=\linewidth]{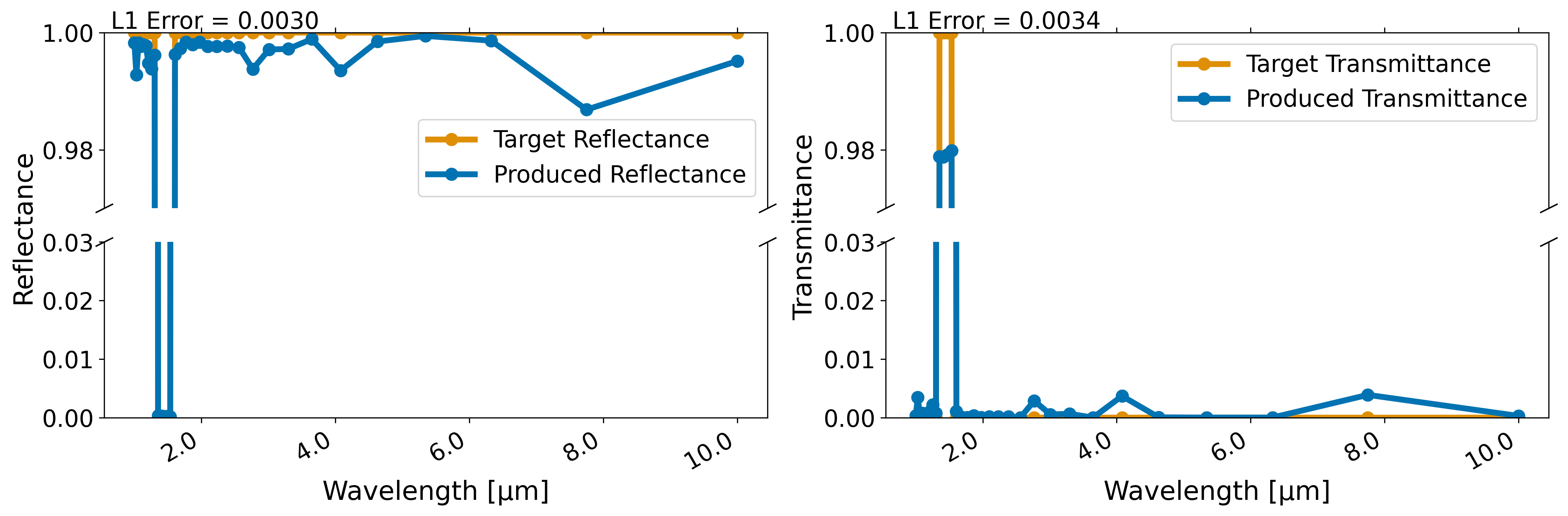}
\caption{Spectral characteristics with an 1550nm exemplary filter designed with NIDN. \label{fig:nidn_2}}
\end{figure}

As exemplary practical applications, we demonstrated two test cases, one of these is displayed in Fig.~\ref{fig:nidn_2} using stacked uniform layers with RCWA. The first showcases creating a 1550nm filter \cite{liao2017long} with NIDN. As can be seen, NIDN is able to produce an almost perfect replica of the spectral characteristics of the desired filter. In the second application demonstrated in the work, results for designing a perfect anti-reflection material are shown. This type of application can, e.g., be relevant to avoid creating satellites which produce specular highlights in astronomical observations \cite{kohler2020astronomy}. The spectral characteristics obtained with NIDN once again were able to match the specified target spectrum extraordinarily well.

\noindent
However, there are still some challenges remaining for this type of approach. In NIDN, currently it is difficult to map the produced permittivities to existing materials as this is a non-differentiable operation. Secondly, parallelization of the utilized simulations is desirable to improve computational efficiency, especially of FDTD. Overall, this line of research seems promising, especially given the low requirements. No training data is needed, limitations on the conceivable structures are only due to chosen Maxwell equations solver and the approach is agnostic to both, the concrete application (e.g. solar sails or filters) and modular with regards to the used Maxwell solver (FDTD, RCWA, etc.) as long as it can be implemented in a differentiable way. 

\subsubsection{Inverse Design of 3D-printable Lattice Materials}

The approach can also be applied to other types of materials.
In an ongoing project \cite{dold2023lattice}, a similar inverse design approach is being developed for 3D-printable lattice materials.
In this case, a differentiable finite element solver is implemented to predict the mechanical properties of a lattice, and the lattice structure is adjusted using gradient information to optimize for certain material properties like in-plane stiffness and Poisson's ratio.

More precisely, we represent lattices as mathematical graphs: edges represent lattice beams and nodes represent the spatial locations where beams cross. 
Hence, a lattice $\mathcal{L}$ can be described as a collection of nodes $\mathcal{N}$ and edges $\mathcal{E}$, $\mathcal{L} = \left(\mathcal{N}, \mathcal{E}\right)$.
In addition, both nodes and edges can be assigned attributes $\mathcal{N}_\text{A}$ and $\mathcal{E}_\text{A}$, respectively.
For instance, nodes are by default characterized by their spatial location, while edges can hold information such as beam cross-area, beam shape and Young's modulus.
On this graph structure, we can define functions $F(\mathcal{L}, \mathcal{N}_\text{A}, \mathcal{E}_\text{A}$) that take the graph and its attributes as input and return its material properties, e.g., effective in-plane stiffness, Poisson's ratio or the deformation under certain loading conditions.
\begin{figure}[tb]
    \centering
    \includegraphics[width=0.95\linewidth]{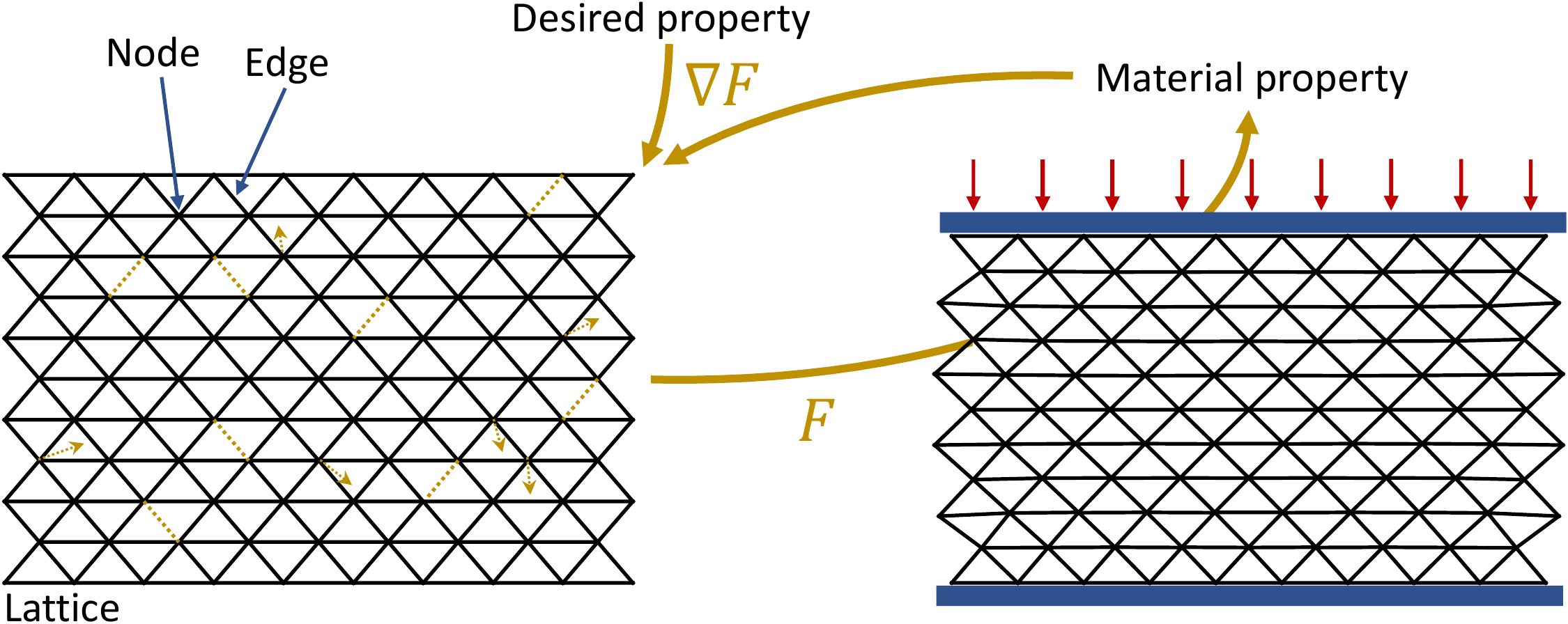}
    \caption{We represent a lattice as a graph with nodes and edges. Inverse design is enabled through a differentiable forward function $F$ that predicts material properties (e.g. the response to loading the lattice from above and below, red arrows indicate loading) from the graph structure alone. Gradient information $\nabla F$ is used to adjust elements of the graph, i.e., adjust the lattice structure by changing node positions and removing or adding edges (yellow dashed lines) with the goal of designing a lattice with desired material properties.}\label{fig:lattice_design}
\end{figure}

To directly work on the graph, $F$ implements functions using message passing which -- at least in most modern approaches -- generalizes the well-known convolution operator from images to graphs \cite{welling2016semi}. 
We found that direct stiffness methods as well as loading experiments can be formulated in a fully differentiable way using this technique.
In this case, $F$ calculates lattice properties exactly while its gradient provides sufficient information to iteratively adjust both node and edge attributes until a lattice structure with the desired target property has been found (\cref{fig:lattice_design}) -- completely without having to train on data.
To change the beam structure of the lattice, we assign each edge $e$ a masking value $m_e$, from which the existence of a beam is derived using thresholding, $\vartheta_e = 1$ if $m_e > 0$ and $\vartheta_e = 0$ otherwise, i.e., $\vartheta_e = \theta(m_e)$ with $\theta(\cdot)$ being the Heaviside function.
This way, contributions from edge $e$ are either added ($\vartheta_e = 1$) or removed ($\vartheta_e = 0$) from $F$ by multiplying edge messages with this term, allowing us to change the beam connectivity via gradient information for $m_e$.
However, gradient-based optimization of $m_e$ is incredibly slow since the derivative of $\theta(\cdot)$ is a Dirac delta distribution. This is solved by using the surrogate gradient method (see \cite{neftci2019surrogate} for details), enabling geometric changes (moving nodes, removing or adding edges) during inverse design.
Finally, instead of a finite element solver, the function $F$ used for inverse design can also be a neural network -- in this case a graph neural network, an architecture that generalizes ANNs to graph-structured input -- trained to predict material properties from lattice graphs. 

Materials used to build future space infrastructure, e.g., for habitats on other celestial bodies, will require very unique, environment-specific properties with the constraint of being easy to process and shape.
The geometric freedom of 3D-printed lattice materials allows for a huge range of customizable properties while satisfying this constraint.
Thus, it is not surprising that at ESA, research on 3D printing lattices from a variety of space-relevant polymers and metals has been growing over the past decade \cite{makaya2022towards,mitchell2018additive}.
Computational inverse design approaches as presented here will greatly assist the design of novel lattice materials for future space applications -- especially by extending the design space to unintuitive structures such as irregular lattices as found in nature.

%% file: sections/onboardAI.tex
\section{Onboard Machine Learning}
Many of the emerging trends using machine learning for space applications, in particular all the differentiable intelligence methods introduced in the previous sections, are greatly empowered if inference and training can be performed directly onboard the spacecraft and not on the ground.
The use of machine learning (ML) methods onboard spacecraft (pioneered in 2003 by JPL's Earth observation mission EO-1 \cite{ungar2003overview}) is receiving renewed attention as the next generation of spacecraft hardware is targeted at supporting specifically these methods and their applications \cite{PhiSat1, mateo2021towards, ruuvzivcka2022ravaen, del2021board}. With launch costs decreasing and the consequent increase in number of payloads sent -- especially to low-Earth orbit (LEO) \cite{lemmens2020esa,jones2018recent} -- operating these spacecraft has become more accessible and commercially viable. The first applications of ML methods for inference have already been demonstrated and, increasingly, efforts are being undertaken to move also the training on board. This section will first explore the motivations for performing ML onboard a spacecraft. Next, the transition from \lq\lq classical\rq\rq spacecraft hardware to current and next-generation hardware \cite{furano2020towards} is presented in the context of current spacecraft hardware trends. After that, a few examples are given to showcase actual in-orbit demonstrations of ML applications, and a particular emphasis is given to ``The OPS-SAT Case'', a competition to develop models suitable to inference onboard the OPS-SAT satellite from the European Space Agency. We conclude by touching upon preliminary results from our work on onboard and decentralized training in space.

\subsection{Why Have Machine Learning on Board?}\label{subsec: onboardMLBackground} 
From an ML perspective, the space environment is particularly challenging if the methods developed must run onboard the spacecraft \cite{furano2020towards}. 
Hardware that is typically used for advanced ML techniques -- such as graphics cards -- has high power consumption, creates excess heat and is usually not radiation-hardened. All three of these aspects are detrimental to the viability of ML in space, with heat dissipation being a critical problem for spacecraft in general \cite{hengeveld2010review}. 
Power is usually only available in the form of solar power -- the most common power source in Earth orbit -- and most satellites are designed with strict power constraints (in the case of small satellites often in the range of few Watts \cite{furano2020towards, pitonak2022cloudsatnet}). 
Finally, typical spacecraft hardware has to be at least radiation-tolerant and, if orbits beyond LEO are considered, radiation-hardened.
With all these issues, it is only natural to question why there is such a strong push to bring heavy ML computations to space in the first place.
A satisfactory answer is revealed, once again, by considering the constraints of operating spacecraft, most prominently communication: downlinks to Earth -- especially in LEO -- depend on short communication windows with ground stations. Thus, depending on the orbit and number of ground stations, a satellite often has only a few minutes to communicate with the ground on a day. These limited windows introduce a high latency between observations made by a satellite and when the data is available and processed on ground. 
Depending on the application, this high latency is a critical issue. For example, the detection of natural disasters, such as wildfires or volcano eruptions \cite{mateo2021towards, del2021board,ruuvzivcka2022ravaen}, in practice requires a rapid response, and minimizing delays is imperative.
Secondly, downlinks to Earth are always limited in terms of bandwidth. For instance, the 6U CubeSat HYPSO-1 (HYPer-spectral Smallsat for ocean Observation) is equipped with 1 Mbps downlink in the S-band \cite{danielsen2021self}, while the 6U CubeSat OPS-SAT is equipped with an X-band 50 Mbps downlink. It is thus often not practical to transmit all the data captured by the onboard sensors to the ground segment and wait for processing to be performed on the ground. At the beginning of our millenia, the satellite EO-1 pioneered the use of onboard ML \cite{ungar2003overview} showcasing the potential advantages in 
autonomous detection and response to science events occurring on the Earth. More recently, the work by Bradley and Brandon \cite{OrbitalEdge} showcases to what extent leveraging onboard processing could ensure better scalability and reduced need for additional ground stations while keeping the same capability to download data for future nanosatellite constellations. 

\subsection{Hardware for Onboard Machine Learning}
\label{subsec: OnboardHW}
The improvement in hardware technology is fundamental to enabling ML onboard spacecraft. Size, weight and power constraints have a significant impact on the choice of hardware and the algorithms that can be implemented on board. Given that some ML methods, such as those based on neural networks, are computationally intensive, the choice of energy-efficient hardware becomes fundamental to enabling the use of ML onboard \cite{pitonak2022cloudsatnet}. 
Given the strict safety requirements and the limited need for computing power, the design of space components has been much more geared towards resilience against harsh thermal cycles, electrostatic discharges, radiation and other phenomena typical of the space environment, rather than ensuring high performance and energy efficiency \cite{lentaris2018high, furano2018roadmap}.
Because of that, numerous works have been investigating the use of commercial off-the-shelf (COTS) devices for non-mission-critical space applications, which generally offer more convenient trade-offs in terms of power consumption, performance, cost and mass \cite{furano2020towards, UnibapSpaceCloud, PhiSat1, OPSSATOnboardAI, danielsen2021self}.

For neural network inference, such devices include AI processors, such as the Intel\textsuperscript{\textregistered} Movidius\textsuperscript{TM} Myriad\textsuperscript{TM} 2 used in the ${\Phi}$-Sat-1 mission \cite{PhiSat1}, field-programmable gate arrays (FPGAs) and system-on-a-chip FPGAs used onboard the OPS-SAT \cite{OPSSATOnboardAI} and HYPSO-1 \cite{danielsen2021self, pitonak2022cloudsatnet} missions, and graphics processing units (GPUs) \cite{OrbitalEdge, UnibapSpaceCloud}. 
To assess the usability of COTS hardware in space, some devices have been tested \cite{UnibapSpaceCloud, furano2020towards} under radiation, demonstrating sufficient resistance for short-term non-critical missions.
ESA's ACT is investigating novel hardware solutions that might enable future applications. 
In the frame of \emph{Onboard \& Distributed Learning} research, presented in Section \ref{subsec: OnboardLearning}, the onboard availability of space qualified GPUs is discussed as an enabler for the distributed training of machine learning models \cite{UnibapSpaceCloud}. 
In the frame of the research on \emph{Neuromorphic sensing \& processing}, described in Section \ref{sec:Neuromorphic}, neuromorphic hardware is suggested as an alternative to conventional von Neumann computing architectures to enable complex computations under extreme low-power constraints.

\subsection{Onboard Inference}

As mentioned in Section \ref{subsec: onboardMLBackground}, numerous researchers have been investigating the possibility of using AI onboard satellites to extract actionable information with reduced latency. 
In 2020, the $\Phi$-Sat-1 mission demonstrated the possibility of discarding cloud-covered images onboard the satellite through inference enabled by a convolutional neural network (CNN) running on a COTS processor \cite{PhiSat1}.
Similarly, the 6U CubeSat HYPSO-1 exploits self-organizing maps to perform onboard real-time detection of harmful algal bloom \cite{danielsen2021self}. 
Recent work \cite{OPSSATOnboardAI} describes the use of ML onboard the 6U ESA OPS-SAT CubeSat, which is an in-orbit laboratory that enables the testing of new software and mission control technologies. 
In particular, \cite{OPSSATOnboardAI} demonstrates the ability of modern AI techniques to detect and discard \lq\lq unwanted\rq\rq\ images for the purpose of reducing the amount of data to download. 

\subsubsection{The OPS-SAT case}
A recent event dedicated to advancing onboard machine learning was a competition organized via ESA's Kelvins platform\footnote{\url{https://kelvins.esa.int/}. Accessed 23/11/22} called \lq\lq The OPS-SAT case\rq\rq\cite{derksenfew}. 
In the \lq\lq The OPS-SAT case\rq\rq, the competitors were given a (quantized) neural model  -- an \textit{EfficientNet-lite-0} \cite{tan2019efficientnet} -- that was tested and passed all the requirements for inference onboard the European Space Agency's OPS-SAT satellite. 
The challengeconsisted in tuning the parameters of the model to enable it to predict one of eight classes for cropped patches coming directly from the spacecraft's raw imaging sensor data. The most performant models trained in the context of the competition were selected to run onboard OPS-SAT during a dedicated flight campaign.


\begin{figure}
    \centering
    \includegraphics[width=0.8\linewidth]{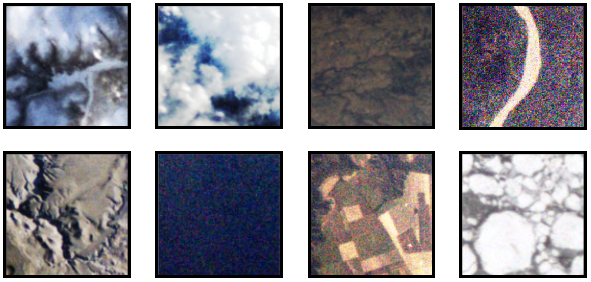}
    \caption{Examples of post-processed patches for each class (Snow, Cloud, Natural, River, Mountain, Water, Agricultural, Ice) in the ``The OPS-SAT case'' competition.}\label{fig: opssatPatches}
\end{figure}

Three main aspects were targeted by the competition setup. The primary one was the use of a limited number of labeled patches. This fact is of major importance to future onboard learning applications where labels are expected to be scarce because of the high effort and cost of labeling Earth observation images \cite{gomez2021msmatch} as well as the need to reduce the operation time required for labeling when a novel sensor is used \cite{derksenfew}.
The second aspect concerned the use of raw satellite images. The need to post-process images increases the number of operations to be performed onboard and consequently the energy and the total time to process one image.
Finally, the use of quantization-aware training to enable the use of quantized models (16-bit floating point). Quantization is, in fact, fundamental to decreasing the model size in order for it to match the satellite uplink requirements \cite{derksenfew}.  

\begin{figure}
    \centering
    \includegraphics[width=0.8\linewidth]{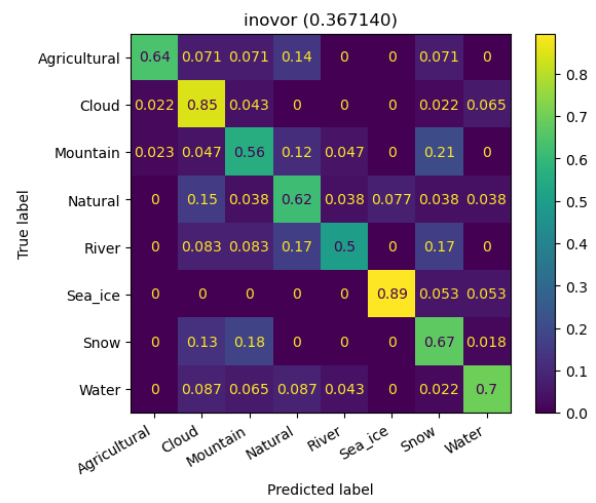}
    \caption{Confusion matrix of the best-performing submission on the evaluation dataset for ``The OPS-SAT Case''.}\label{fig: OPSSATWinnerCOM}
\end{figure}

To learn the model weights, a dataset was manually created containing patches of $200 \times 200$ pixels produced cropping the raw onboard sensor data. The dataset was then split into \textit{training} and \textit{evaluation}, of which only the first one was released to the competitors. 
The \textit{training} dataset \cite{OPSSATDataset} contained $10$ labeled patches per class and $23$ unprocessed original images, while the evaluation dataset consisted of $588$ labeled patches. 
Given the different number of available evaluation patches for each of the classes (Agricultural: 36, Cloud: 114, Mountain: 99, Natural: 58, River: 31, Ice: 37, Snow: 106, Water: 107), the evaluation dataset was unbalanced. 
Because of that, model accuracy was not an appropriate metric for evaluating the competition results since it would not have significantly penalised submissions underscoring on the  \lq\lq River \rq\rq and \lq Ice\rq\rq classes. 
Hence, the metric used for the competition was devised as follows:
\begin{figure}
    \centering
    \includegraphics[width=0.8\linewidth]{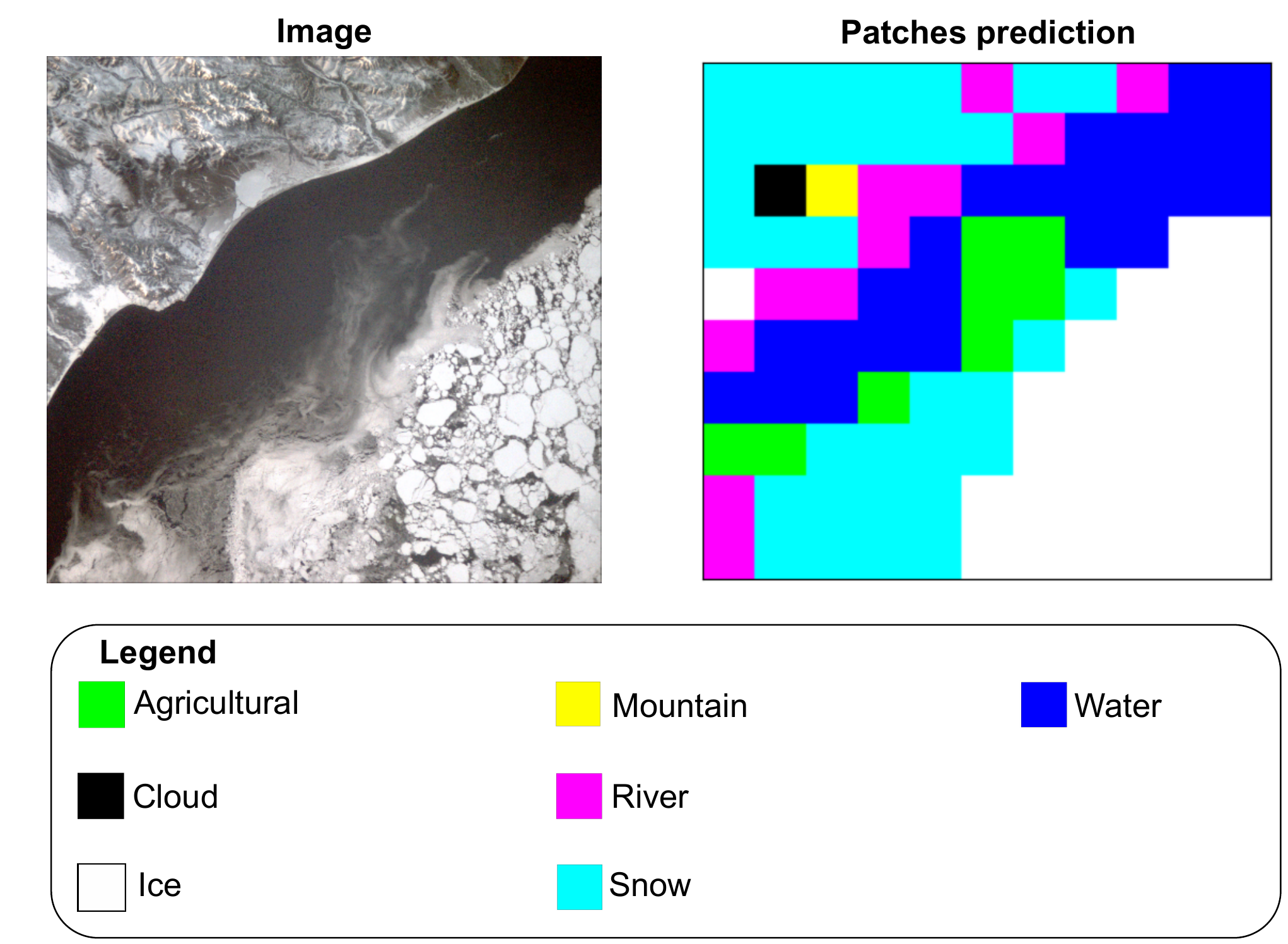}
    \caption{(Left) A post-processed image from the camera onboard OPS-SAT. (Right) Classification of the correspondent $200 \times 200$ patches performed by the competition baseline model.}\label{fig: opssatPatchesClassification}
\end{figure}

\begin{equation}
    \mathcal{L} = 1 - \kappa
\end{equation}
\noindent
where $\kappa$ is \textit{Cohen's kappa} coefficient, which is used extensively for the evaluation of classification performance on unbalanced datasets in remote sensing scenarios \cite{musial2022comparison,deshpande2021historical}. 
To provide a baseline for the competition, we trained an \textit{EfficientNet-lite-0} model by using \textit{MSMatch} \cite{gomez2021msmatch}, which is a machine learning approach developed by the ACT for the classification of MultiSpectral and RGB remote sensing data. Fig.\ref{fig: opssatPatchesClassification} shows the results of the competition baseline model on reconstructing images from the \emph{training} set. 
The $41$ teams that participated in the three-month-long competition produced a total of $891$ submissions. The highest score on the evaluation dataset was $\mathcal{L} = 0.367140$. For reference, the competition baseline model obtained a score of $\mathcal{L} = 0.539694$ and would have ranked $12th$. Fig.\ref{fig: OPSSATWinnerCOM} shows the confusion matrix produced by the leading submission on the evaluation dataset. 
The confusion matrix for the submission from the winning team \emph{inovor} shows that the winning model is capable of distinguishing between the \lq\lq Cloud\rq\rq and \lq\lq Ice\rq\rq classes, which have respectively $85\%$ and $89\%$ true positive rates. The \lq\lq River\rq\rq\ class had the worst true positive rate ($50\%$), with misclassified \lq\lq River\rq\rq patches split equally between the \lq\lq Natural\rq and \lq\lq Snow\rq\rq classes ($17\%$ each). For comparison, only $12.8\%$ of the other classes were misclassified as \lq\lq River\rq\rq. Finally, it should be noted that $21\%$ of \lq\lq Mountain\rq\rq patches were classified as \lq\lq Snow\rq\rq, while $18\%$ of \lq\lq Snow\rq\rq patches were classified as \lq\lq Mountain\rq\rq. This could be due to the presence of common features between the two classes, such as the presence of mountains in the \lq\lq Snow \rq\rq patches or the presence of snow (or other elements that cannot be easily distinguished from snow in an RGB image) in the \lq\lq Mountain\rq\rq\ patches. A dedicated publication authored with the winning teams will contain a detailed analysis of the competition preparation, its results, the machine learning approaches developed and the inference results obtained onboard OPS-SAT.

\subsection{Onboard \& Distributed Learning}\label{subsec: OnboardLearning}
At the time of writing, to the authors' knowledge, no space mission has yet demonstrated end-to-end training of a large machine learning model onboard a spacecraft. There are many reasons for this, including the lack of suitable hardware on most spacecraft and the large amounts of power and data required for training machine learning methods, especially large neural networks. 
As a reference, the training of current large-scale language models can require power on the order of gigawatthours \cite{patterson2021carbon}. However, small and efficient models, such as the EfficientNet architecture \cite{tan2021efficientnetv2}, can be trained even on small edge devices, such as those described in the previous section. 
Although power consumption is still a relevant factor, onboard training of large machine learning models is entering the realm of possibility. In fact, there are several factors that make this relevant and interesting from a practical point of view. 
Similar to inference, performing training onboard (as opposed to doing so on the ground and then updating the ML model parameters), reduces the communication bandwidth required while potentially contributing to the mission autonomy significantly.
Depending on the location of the spacecraft, sending to and receiving data from the ground may even be completely infeasible. From that perspective onboard training may become essential for mission scenarios where autonomy is of paramount importance. 
For instance, this is the case for deep-space missions where communication becomes increasingly difficult as the spacecraft moves further away from Earth. 
Furthermore, assuming learning can be performed onboard, the application of decentralized learning approaches \cite{mcmahan2017communication} are enabled for fleets or constellations of spacecraft. 
This adds a requirement on the existence of inter-satellite links, one that is in synergy with proposals advocating for advanced optical inter-satellite communications (see for example \cite{kaur2019analysis}).

Distributed learning has lately generated an increasing interest due to the combination of the number of commercial activities around satellite constellations \cite{massey2020challenge} and the broader availability of hardware capable of running machine learning algorithms in space\cite{spacecloud}. In particular, the potential of federated learning (a distributed learning paradigm) is being explored for joint training of satellite constellations \cite{razmi2022board,matthiesen2022federated}.

Together with several collaborators, the ACT is supporting this effort with the design of an open-source software package called PASEOS (PAseos Simulates the Environment for Operating multiple Spacecraft) \footnote{\url{https://github.com/aidotse/PASEOS} Accessed: 2022-10-28}. PASEOS aims to model the operational constraints of spacecraft with respect to decentralized learning and similar activities. It operates in a fully decentralised manner, where each device (or `actor' in PASEOS terminology) runs an instance of the PASEOS software that provides it with a model of the simulated spacecraft's power, thermal, radiation, communication and orbital constraints. The motivation behind the software is that, in comparison to the well-explored decentralized and federated learning approaches on Earth \cite{lim2020federated}, the operational constraints for training ML models onboard spacecraft are different and unique. Thus, in order to realistically explore these applications, it is important to find efficient and feasible solutions to these challenges by first specifying the constraints relevant to space.